\documentclass[
]{ceurart}
\usepackage{listings}
\usepackage{url}
\begin{document}

\copyrightyear{2023}
\copyrightclause{Copyright for this paper by its authors.
  Use permitted under Creative Commons License Attribution 4.0
  International (CC BY 4.0).}

\conference{IberLEF 2023, September 2023, Jaén, Spain}

\title{Generative AI Text Classification using Ensemble LLM Approaches}


\author[1]{Harika Abburi}[%
email=abharika@deloitte.com,
]
\cormark[1]

\author[2]{Michael Suesserman}[%
email=msuesserman@deloitte.com,
]

\author[1]{Nirmala Pudota}[%
email=npudota@deloitte.com,
]

\author[2]{Balaji Veeramani}[%
email=bveeramani@deloitte.com,
]

\author[2]{Edward Bowen}[%
email=edbowen@deloitte.com,
]


\author[2]{Sanmitra Bhattacharya}[%
email=sanmbhattacharya@deloitte.com,
]

\address[1]{Deloitte \& Touche Assurance \& Enterprise Risk Services India Private Limited, India}
\address[2]{Deloitte \& Touche LLP, USA}

\cortext[1]{Corresponding author.}

\begin{abstract}
Large Language Models (LLMs) have shown impressive performance across a variety of Artificial Intelligence (AI) and natural language processing tasks, such as content creation, report generation, etc. However, unregulated malign application of these models can create undesirable consequences such as generation of fake news, plagiarism, etc. As a result, accurate detection of AI-generated language can be crucial in responsible usage of LLMs. In this work, we explore 1) whether a certain body of text is AI generated or written by human, and 2) attribution of a specific language model in generating a body of text. Texts in both English and Spanish are considered. The datasets used in this study are provided as part of the Automated Text Identification (AuTexTification) shared task. For each of the research objectives stated above, we propose an ensemble neural model that generates probabilities from different pre-trained LLMs which are used as features to a Traditional Machine Learning (TML) classifier following it. For the first task of distinguishing between AI and human generated text, our model ranked in fifth and thirteenth place (with macro $F1$ scores of 0.733 and 0.649) for English and Spanish texts, respectively. For the second task on model attribution, our model ranked in first place with macro $F1$ scores of 0.625 and 0.653 for English and Spanish texts, respectively.
\end{abstract}

\begin{keywords}
  generative AI \sep
  text classification \sep
  large language models \sep
  ensemble
\end{keywords}

\maketitle

\section{Introduction}
Rapid advances in the capabilities of LLMs, and their ease of use in generating sophisticated and coherent content is leading to the production of AI-generated content at scale. Some of the applications of LLMs are in AI-assisted writing \cite{hutson2021robo}, medical question answering  \cite{kung2023performance, wang2021systematic} and a wide range of tasks in the financial services industry \cite{wu2023bloomberggpt} and legal domain \cite{sun2023short}. Foundational models such as OpenAI’s GPT-3 \cite{brown2020language}, Meta’s OPT \cite{zhang2022opt}, and Big Science’s BLOOM \cite{scao2022bloom} can generate such sophisticated content with basic text prompts that it is often challenging to manually discern between human and AI-generated text. While these models demonstrate the ability to understand the context and generate coherent human-like responses, they do not have a true understanding of what they are producing \cite{li2023ethics,turpin2023language}. This could potentially lead to adverse consequences when used in downstream applications. For example, consider an application of a LLM used for summarizing a medicinal drug data-sheet that inadvertently produces wrong dosage information. Generating plausible but false content (referred to as \emph{hallucination} \cite{bang2023multitask, ji2023survey}), may inadvertently help propagate misinformation, false narratives, fake news, and spam. Given the pace of LLM adoption by the general public, and the rate of dissemination of information across the globe, widespread  misinformation propagation is an imminent risk that both individuals and organizations will have to  deal with in the near future \cite{ali2023chatgpt, weidinger2021ethical}. An understanding of the source of the authored content -- whether by an AI system or a human -- would allow one to appropriately use the content in downstream applications with suitable oversight. In the case of AI-generated content, the knowledge of the source LLM would allow one to watch out for potential known biases and limitations associated with that model. Given the risks associated with unchecked and unregulated adoption of generative AI models, it is imperative to develop approaches to detect AI-generated content, and identify the source of the AI-generated content.

Motivated by these challenges, automatic detection of AI-generated text has become an active area of research. Recent work such as DetectGPT \cite{mitchell2023detectgpt} generates minor perturbations of a passage using a generic pre-trained Text-to-Text Transfer Transformer (T5) model, and then compares the log probability of the original sample with each perturbed sample to determine if it is AI generated. Another approach \cite{li2022artificial} developed a model for a generative multi-class AI detection challenge involving Russian language text using Decoding- enhanced BERT with disentangled attention (DeBERTa) as a pre-trained language model for classification  and an ensemble approach is proposed by \cite{maloyan2022dialog} for binary classification. Mitrovic \emph{et al.} \cite{mitrovic2023chatgpt} developed a fine-tuned transformer-based approach to distinguish between human and ChatGTP generated text, with addition of SHapely Additive exPlanations (SHAP) values for model explainability. Statistical detection methods have also been applied for detection of generative AI text, such as the Giant Language model Test Room (GLTR) approach developed by researchers \cite{gehrmann2019gltr}. Research involving repeated higher-order n-grams found that they occur more often in AI generated text as compared to human generated text \cite{galle2021unsupervised}. Using an ensemble of classifiers, higher-order n-grams can be used to help distinguish between human and AI generated text.

To boost this area of research further, the Automated Text Identification (AuTexTification) \cite{autextification} shared task, part of IberLEF 2023 \cite{overviewIberLEF2023}, put forth two tasks, each covering both English and Spanish language texts: 1. \emph{Human or generated:} determine whether a given input text has been automatically generated or not -- a binary classification task with two classes `human' or `AI' 2. \emph{Model attribution:} for automatically generated text, determine which one of six different text generation AI models was the source -- a multi-class classification task with six classes, where each class represents a text generation model. For each of these tasks, we propose an ensemble classifier, where the probabilities generated from various state-of-the-art LLMs are used as input feature vectors to TML models to produce final predictions. Our experiments show multiple instances of the proposed framework outperforming several baselines using established evaluation metrics.
\section{Dataset}
The data for all the subtasks and languages, is provided by the AuTexTification shared task organizers \cite{autextification}. For Task 1, \emph{Human or generated}, the data consists of texts from five domains. During the training phase of the shared task the domains are not disclosed to the participants. For Task 1, for both English and Spanish, data from three domains is provided as training data, and data from two new domains is provided in the test set. For Task 2, \emph{Model attribution}, also the data comes from five different domains, but the same domains are in train and test splits. This data is evenly distributed into six class (A, B, C, D, E, and F), where each class represents a text generation model. An interesting point to note here is that the six text generation models are of increasing number of neural parameters, ranging from 2B to 175B. The motivation here is to emulate realistic AI text detection approaches which should be  versatile enough to detect a diverse set of text generation models and writing styles. More details about the data can be found in the AuTextification overview paper \cite{autextification}.

\section{Proposed Ensemble Approach}
\label{approach}
In this section, we describe our approaches for detection and classification of generative AI text.

\subsection{Models}
We explored various state-of-the-art large language models \cite{wolf2020transformers} such as Bidirectional Encoder Representations from Transformers (BERT), DeBERTa, Robustly optimized BERT approach (RoBERTa), and cross-lingual language model RoBERTa (XLM-RoBERTa) along with their variants. Since the datasets are different for each task and language, and the same set of the models will not fit across them, we fine-tuned different models for different subtasks and languages and pick the best models based on validation data. Table \ref{models} lists the different models that we explored for the different subtasks: Task 1 in English (Binary-English), Task 1 in Spanish (Binary-Spanish), Task 2 in English (Multiclass-English) and Task 2 in Spanish (Multiclass-Spanish). We also explored different TML models such as Linear SVC \cite{sulaiman2020question}, Error-Correcting Output Codes (ECOC) \cite{liu2023novel}, OneVsRest \cite{hong2008probabilistic} and Voting classifier which includes Logistic Regression (LR), Random Forest (RF), Gaussian Naive Bayes (NB), Support Vector machines (SVC) \cite{mahabub2020robust} in our approach and the best model for each subtask is shown in results section.

\begin{table*}[h]
\caption{Models explored for different tasks}\label{models}
\begin{center}
\begin{tabular}{|p{1.5cm}|p{11cm}|}
\hline
\textbf{Task} &  \textbf{Large language models}\\
\hline
Binary-English & deberta-large \cite{he2021deberta}, xlm-r-100langs-bert-base-nli-stsb-mean-tokens \cite{reimers-2019-sentence-bert}, roberta-base-openai-detector \cite{solaiman2019release}, xlm-roberta-large-xnli-anli, roberta-large\\
\hline
Binary-Spanish & bertin-roberta-base-spanish \cite{BERTIN}, MarIA \cite{MARIA}, sentence\_similarity\_spanish\_es, xlm-roberta-large-xnli-anli, xlm-roberta-large-finetuned-conll02-spanish \cite{conneau2019unsupervised} \\
\hline
Multiclass-English &  xlm-roberta-large-finetuned-conll03-english, scibert\_scivocab\_cased \cite{beltagy-etal-2019-scibert}, deberta-base, roberta-large \cite{DBLP:journals/corr/abs-1907-11692}, longformer-base-4096 \cite{Beltagy2020Longformer}, bert-large-uncased-whole-word-masking-finetuned-squad \cite{DBLP:journals/corr/abs-1810-04805} \\
\hline
Multiclass-Spanish & xlm-roberta-large-finetuned-conll03-english, MarIA, sentence\_similarity\_spanish\_es, bert-base-multilingual-cased-finetuned-conll03-spanish, roberta-large \\
\hline
\end{tabular}
\end{center}
\end{table*}
\subsection{Proposed Ensemble Approach}
Each input text is passed through variants of the pre-trained large language models such as DeBERTa (D), XLM-RoBERTa (X), RoBERTa (R), BERT (B), etc. During the model training phase, these models are fine-tuned on the training data. For inference and testing, each of these models can independently generate classification probabilities (P), namely $P^{D}$, $P^{X}$, $P^{R}$, $P^{B}$, etc. These probabilities are concatenated ($P^{C}$) or averaged ($P^{A}$), and the output is passed as a feature vector to train TML models to produce final predictions.

\section{Experiments}
This section provides the experimental evaluation of the proposed methods. For both the tasks we report results on accuracy ($Acc$), macro F1 score ($F_{macro}$), precision ($Prec$) and recall ($Rec$). 

\subsection{Implementation Details}
We set aside 20\% from the training data for validation. For the testing phase, the validation set is merged with the training set. All the results are reported on testing data. The hyper-parameters are used for model fine-tuning are shown in Table \ref{tab:hyper}.


\begin{table}[h]
\caption{Hyper-parameters for all the subtasks}
\label{tab:hyper}
\begin{center}
\begin{tabular}{|c|c|}
\hline
\textbf{Parameter} & \textbf{value} \\
\hline
Batch size&128\\
\hline
Learning rate & 0.00003\\
\hline
Maximum sequence length & 128\\
\hline
Epochs& 10 for task1 and 20 for task2\\
\hline
\end{tabular}
\label{tab:my_label}
\end{center}
\end{table}

\begin{table}[h]
\caption{Results for the Binary-English task}
\label{tab:Binaryenglish}
\begin{center}
\begin{tabular}{|l|c|c|c|c|}
\hline
\textbf{Model} &  \textbf{$Acc$} & \textbf{$F_{macro}$} & \textbf{$Prec$}  &\textbf{$Rec$} \\
\hline
\hline
deberta-large&0.620&0.546&0.783&0.610\\
\hline
xlm-r-100langs-bert-base-nli-stsb-mean-tokens&0.647&0.592&0.782&0.639\\
\hline
roberta-base-openai-detector&0.679&0.636&0.805&0.671\\
\hline
xlm-roberta-large-xnli-anli&0.618&0.543&0.782&0.608\\
\hline
roberta-large&0.623&0.551&0.784&0.613\\
\hline
Ensemble with Voting classifier ($P^{C}$ as a input feature
)&\textbf{0.751}&\textbf{0.733}&\textbf{0.826}&\textbf{0.745}\\
\hline
\end{tabular}
\end{center}
\end{table}

\subsection{Results}
For each task and language, we submitted three runs to the leaderboard (team name \emph{Drocks}). These runs correspond to the most promising approaches on the validation data. In this paper, we show only the top run results for each of the tasks. The complete leaderboard is available at \emph{https://sites.google.com/view/autextification/results} \cite{autextification}. The results on test data for Binary-English and Binary-Spanish are shown in Tables \ref{tab:Binaryenglish} and \ref{tab:Binaryspanish}, respectively. With the concatenated feature vector ($P^{C}$) as an input, a voting classifier results in $F_{macro}$ score of 73.3 on Binary-English data, whereas a onevsrest classifier outperforms other methods with $F_{macro}$ score of 64.9 on Binary-Spanish data.

\begin{table}[!t]
\caption{Results for the Binary-Spanish task}
\label{tab:Binaryspanish}
\begin{center}
\begin{tabular}{|l|c|c|c|c|}
\hline
\textbf{Model} &  \textbf{$Acc$} & \textbf{$F_{macro}$} & \textbf{$Prec$}  &\textbf{$Rec$} \\
\hline
\hline
bertin-roberta-base-spanish&0.698&0.633&0.798&0.661\\
\hline
MarIA&0.690&0.629&0.791&0.652\\
\hline
sentence\_similarity\_spanish\_es&0.651&0.560&0.786&0.607\\
\hline
xlm-roberta-large-xnli-anli&0.633&0.526&0.788&0.587\\
\hline
xlm-roberta-large-finetuned-conll02-spanish&0.637&0.533&0.787&0.591\\
\hline
Ensemble with OneVsRest classifier ($P^{C}$ as a input feature
)&\textbf{0.704}&\textbf{0.649}&\textbf{0.805}&\textbf{0.667}\\
\hline
\end{tabular}
\end{center}
\end{table}

\begin{table}[!t]
\caption{Results for the Multi-English task}
\label{tab:multienglish}
\begin{center}
\begin{tabular}{|l|c|c|c|c|}
\hline
\textbf{Model} & \textbf{$Acc$} &\textbf{$F_{macro}$} &\textbf{$Prec$}  &\textbf{$Rec$} \\
\hline
\hline
xlm-roberta-large-finetuned-conll03-english&0.598&0.593&0.618&0.594\\
\hline
scibert\_scivocab\_cased&0.578&0.576&0.590&0.575\\
\hline
deberta-base&0.564&0.558&0.602&0.558\\
\hline
roberta-large&0.581&0.568&0.611&0.574\\
\hline
longformer-base-4096&0.586&0.582&0.600&0.582\\
\hline
bert-large-uncased-whole-word-masking-finetuned-squad&0.581&0.581&0.597&0.579\\
\hline
Ensemble with ECOC classifier ($P^{C}$ as a input feature
) &\textbf{0.624}&\textbf{0.625}&\textbf{0.649}&\textbf{0.621}\\
\hline
\end{tabular}
\end{center}
\end{table}

\begin{table}[!t]
\caption{Results for the Multi-Spanish task}
\label{tab:multispanish}
\begin{center}
\begin{tabular}{|l|c|c|c|c|}
\hline
\textbf{Model} &  \textbf{$Acc$} & \textbf{$F_{macro}$} & \textbf{$Prec$}  &\textbf{$Rec$} \\
\hline
\hline
xlm-roberta-large-finetuned-conll03-english&0.632&0.629&0.661&0.628\\
\hline
MarIA&0.614&0.615&0.630&0.612\\
\hline
sentence\_similarity\_spanish\_es&0.615&0.612&0.640&0.613\\
\hline
bert-base-multilingual-cased-finetuned-conll03-spanish&0.593&0.594&0.599&0.593\\
\hline
roberta-large&0.584&0.584&0.595&0.584\\
\hline
Ensemble with Linear SVC classifier ($P^{A}$ as a input feature
)&\textbf{0.653}&\textbf{0.654}&\textbf{0.679}&\textbf{0.650}\\
\hline
\end{tabular}
\end{center}
\end{table}

Tables \ref{tab:multienglish} and  \ref{tab:multispanish} show the results on test data for Multiclass-English and Multiclass-Spanish, respectively. For the Multiclass-English data, an ECOC classifier on top of the concatenated feature vector ($P^{C}$) outperforms the other approaches with $F_{macro}$ score of 62.5. On the other hand, a linear SVC classifier with averaged feature vector ($P^{A}$) as an input outperforms the other approaches with $F_{macro}$ score of 65.4 for the Multiclass-Spanish data.

\section{Conclusion}
In this paper, we described our submission to the AuTexTification shared task which consists of two tasks on the classification of generative AI content. In our experiments, we found that our proposed ensemble LLM approach is a promising strategy, as our model ranked fifth with a macro $F1$ score of 73.3\% for English and thirteenth with 64.9\% macro $F1$ score for Spanish in the \emph{Human or generated} binary classification task. For the \emph{Model attribution} multiclass classification task, our model ranked in the first place for both English and Spanish, with macro $F1$ scores of 62.5\% and 65.3\%, respectively. While our approach shows promising results for the \emph{Model attribution} task, further exploration is needed to enhance and tune our models for \emph{Human or generated} binary classification task.

\bibliography{references}

\begin{thebibliography}{37}
\expandafter\ifx\csname natexlab\endcsname\relax\def\natexlab#1{#1}\fi
\providecommand{\url}[1]{\texttt{#1}}
\providecommand{\href}[2]{#2}
\providecommand{\path}[1]{#1}
\providecommand{\DOIprefix}{doi:}
\providecommand{\ArXivprefix}{arXiv:}
\providecommand{\URLprefix}{URL: }
\providecommand{\Pubmedprefix}{pmid:}
\providecommand{\doi}[1]{\href{http://dx.doi.org/#1}{\path{#1}}}
\providecommand{\Pubmed}[1]{\href{pmid:#1}{\path{#1}}}
\providecommand{\bibinfo}[2]{#2}
\ifx\xfnm\relax \def\xfnm[#1]{\unskip,\space#1}\fi
\bibitem[{Hutson(2021)}]{hutson2021robo}
\bibinfo{author}{M.~Hutson},
\newblock \bibinfo{title}{Robo-writers: the rise and risks of
  language-generating ai},
\newblock \bibinfo{journal}{Nature} \bibinfo{volume}{591}
  (\bibinfo{year}{2021}) \bibinfo{pages}{22--25}.
\bibitem[{Kung et~al.(2023)Kung, Cheatham, Medenilla, Sillos, De~Leon,
  Elepa{\~n}o, Madriaga, Aggabao, Diaz-Candido, Maningo
  et~al.}]{kung2023performance}
\bibinfo{author}{T.~H. Kung}, \bibinfo{author}{M.~Cheatham},
  \bibinfo{author}{A.~Medenilla}, \bibinfo{author}{C.~Sillos},
  \bibinfo{author}{L.~De~Leon}, \bibinfo{author}{C.~Elepa{\~n}o},
  \bibinfo{author}{M.~Madriaga}, \bibinfo{author}{R.~Aggabao},
  \bibinfo{author}{G.~Diaz-Candido}, \bibinfo{author}{J.~Maningo}, et~al.,
\newblock \bibinfo{title}{Performance of chatgpt on usmle: Potential for
  ai-assisted medical education using large language models},
\newblock \bibinfo{journal}{PLoS digital health} \bibinfo{volume}{2}
  (\bibinfo{year}{2023}) \bibinfo{pages}{e0000198}.
\bibitem[{Wang et~al.(2021)Wang, Wang, Yu, Yang, Walker, and
  Mostafa}]{wang2021systematic}
\bibinfo{author}{M.~Wang}, \bibinfo{author}{M.~Wang}, \bibinfo{author}{F.~Yu},
  \bibinfo{author}{Y.~Yang}, \bibinfo{author}{J.~Walker},
  \bibinfo{author}{J.~Mostafa},
\newblock \bibinfo{title}{A systematic review of automatic text summarization
  for biomedical literature and ehrs},
\newblock \bibinfo{journal}{Journal of the American Medical Informatics
  Association} \bibinfo{volume}{28} (\bibinfo{year}{2021})
  \bibinfo{pages}{2287--2297}.
\bibitem[{Wu et~al.(2023)Wu, Irsoy, Lu, Dabravolski, Dredze, Gehrmann,
  Kambadur, Rosenberg, and Mann}]{wu2023bloomberggpt}
\bibinfo{author}{S.~Wu}, \bibinfo{author}{O.~Irsoy}, \bibinfo{author}{S.~Lu},
  \bibinfo{author}{V.~Dabravolski}, \bibinfo{author}{M.~Dredze},
  \bibinfo{author}{S.~Gehrmann}, \bibinfo{author}{P.~Kambadur},
  \bibinfo{author}{D.~Rosenberg}, \bibinfo{author}{G.~Mann},
\newblock \bibinfo{title}{Bloomberggpt: A large language model for finance},
\newblock \bibinfo{journal}{arXiv preprint arXiv:2303.17564}
  (\bibinfo{year}{2023}).
\bibitem[{Sun(2023)}]{sun2023short}
\bibinfo{author}{Z.~Sun},
\newblock \bibinfo{title}{A short survey of viewing large language models in
  legal aspect},
\newblock \bibinfo{journal}{arXiv preprint arXiv:2303.09136}
  (\bibinfo{year}{2023}).
\bibitem[{Brown et~al.(2020)Brown, Mann, Ryder, Subbiah, Kaplan, Dhariwal,
  Neelakantan, Shyam, Sastry, Askell et~al.}]{brown2020language}
\bibinfo{author}{T.~Brown}, \bibinfo{author}{B.~Mann},
  \bibinfo{author}{N.~Ryder}, \bibinfo{author}{M.~Subbiah},
  \bibinfo{author}{J.~D. Kaplan}, \bibinfo{author}{P.~Dhariwal},
  \bibinfo{author}{A.~Neelakantan}, \bibinfo{author}{P.~Shyam},
  \bibinfo{author}{G.~Sastry}, \bibinfo{author}{A.~Askell}, et~al.,
\newblock \bibinfo{title}{Language models are few-shot learners},
\newblock \bibinfo{journal}{Advances in neural information processing systems}
  \bibinfo{volume}{33} (\bibinfo{year}{2020}) \bibinfo{pages}{1877--1901}.
\bibitem[{Zhang et~al.(2022)Zhang, Roller, Goyal, Artetxe, Chen, Chen, Dewan,
  Diab, Li, Lin et~al.}]{zhang2022opt}
\bibinfo{author}{S.~Zhang}, \bibinfo{author}{S.~Roller},
  \bibinfo{author}{N.~Goyal}, \bibinfo{author}{M.~Artetxe},
  \bibinfo{author}{M.~Chen}, \bibinfo{author}{S.~Chen},
  \bibinfo{author}{C.~Dewan}, \bibinfo{author}{M.~Diab},
  \bibinfo{author}{X.~Li}, \bibinfo{author}{X.~V. Lin}, et~al.,
\newblock \bibinfo{title}{Opt: Open pre-trained transformer language models},
\newblock \bibinfo{journal}{arXiv preprint arXiv:2205.01068}
  (\bibinfo{year}{2022}).
\bibitem[{Scao et~al.(2022)Scao, Fan, Akiki, Pavlick, Ili{\'c}, Hesslow,
  Castagn{\'e}, Luccioni, Yvon, Gall{\'e} et~al.}]{scao2022bloom}
\bibinfo{author}{T.~L. Scao}, \bibinfo{author}{A.~Fan},
  \bibinfo{author}{C.~Akiki}, \bibinfo{author}{E.~Pavlick},
  \bibinfo{author}{S.~Ili{\'c}}, \bibinfo{author}{D.~Hesslow},
  \bibinfo{author}{R.~Castagn{\'e}}, \bibinfo{author}{A.~S. Luccioni},
  \bibinfo{author}{F.~Yvon}, \bibinfo{author}{M.~Gall{\'e}}, et~al.,
\newblock \bibinfo{title}{Bloom: A 176b-parameter open-access multilingual
  language model},
\newblock \bibinfo{journal}{arXiv preprint arXiv:2211.05100}
  (\bibinfo{year}{2022}).
\bibitem[{Li et~al.(2023)Li, Moon, Purkayastha, Celi, Trivedi, and
  Gichoya}]{li2023ethics}
\bibinfo{author}{H.~Li}, \bibinfo{author}{J.~T. Moon},
  \bibinfo{author}{S.~Purkayastha}, \bibinfo{author}{L.~A. Celi},
  \bibinfo{author}{H.~Trivedi}, \bibinfo{author}{J.~W. Gichoya},
\newblock \bibinfo{title}{Ethics of large language models in medicine and
  medical research},
\newblock \bibinfo{journal}{The Lancet Digital Health}  (\bibinfo{year}{2023}).
\bibitem[{Turpin et~al.(2023)Turpin, Michael, Perez, and
  Bowman}]{turpin2023language}
\bibinfo{author}{M.~Turpin}, \bibinfo{author}{J.~Michael},
  \bibinfo{author}{E.~Perez}, \bibinfo{author}{S.~R. Bowman},
\newblock \bibinfo{title}{Language models don't always say what they think:
  Unfaithful explanations in chain-of-thought prompting},
\newblock \bibinfo{journal}{arXiv preprint arXiv:2305.04388}
  (\bibinfo{year}{2023}).
\bibitem[{Bang et~al.(2023)Bang, Cahyawijaya, Lee, Dai, Su, Wilie, Lovenia, Ji,
  Yu, Chung et~al.}]{bang2023multitask}
\bibinfo{author}{Y.~Bang}, \bibinfo{author}{S.~Cahyawijaya},
  \bibinfo{author}{N.~Lee}, \bibinfo{author}{W.~Dai}, \bibinfo{author}{D.~Su},
  \bibinfo{author}{B.~Wilie}, \bibinfo{author}{H.~Lovenia},
  \bibinfo{author}{Z.~Ji}, \bibinfo{author}{T.~Yu}, \bibinfo{author}{W.~Chung},
  et~al.,
\newblock \bibinfo{title}{A multitask, multilingual, multimodal evaluation of
  chatgpt on reasoning, hallucination, and interactivity},
\newblock \bibinfo{journal}{arXiv preprint arXiv:2302.04023}
  (\bibinfo{year}{2023}).
\bibitem[{Ji et~al.(2023)Ji, Lee, Frieske, Yu, Su, Xu, Ishii, Bang, Madotto,
  and Fung}]{ji2023survey}
\bibinfo{author}{Z.~Ji}, \bibinfo{author}{N.~Lee},
  \bibinfo{author}{R.~Frieske}, \bibinfo{author}{T.~Yu},
  \bibinfo{author}{D.~Su}, \bibinfo{author}{Y.~Xu}, \bibinfo{author}{E.~Ishii},
  \bibinfo{author}{Y.~J. Bang}, \bibinfo{author}{A.~Madotto},
  \bibinfo{author}{P.~Fung},
\newblock \bibinfo{title}{Survey of hallucination in natural language
  generation},
\newblock \bibinfo{journal}{ACM Computing Surveys} \bibinfo{volume}{55}
  (\bibinfo{year}{2023}) \bibinfo{pages}{1--38}.
\bibitem[{Ali et~al.(2023)Ali, Qadir, and Shah}]{ali2023chatgpt}
\bibinfo{author}{H.~Ali}, \bibinfo{author}{J.~Qadir},
  \bibinfo{author}{Z.~Shah},
\newblock \bibinfo{title}{Chatgpt and large language models (llms) in
  healthcare: Opportunities and risks}  (\bibinfo{year}{2023}).
\bibitem[{Weidinger et~al.(2021)Weidinger, Mellor, Rauh, Griffin, Uesato,
  Huang, Cheng, Glaese, Balle, Kasirzadeh et~al.}]{weidinger2021ethical}
\bibinfo{author}{L.~Weidinger}, \bibinfo{author}{J.~Mellor},
  \bibinfo{author}{M.~Rauh}, \bibinfo{author}{C.~Griffin},
  \bibinfo{author}{J.~Uesato}, \bibinfo{author}{P.-S. Huang},
  \bibinfo{author}{M.~Cheng}, \bibinfo{author}{M.~Glaese},
  \bibinfo{author}{B.~Balle}, \bibinfo{author}{A.~Kasirzadeh}, et~al.,
\newblock \bibinfo{title}{Ethical and social risks of harm from language
  models},
\newblock \bibinfo{journal}{arXiv preprint arXiv:2112.04359}
  (\bibinfo{year}{2021}).
\bibitem[{Mitchell et~al.(2023)Mitchell, Lee, Khazatsky, Manning, and
  Finn}]{mitchell2023detectgpt}
\bibinfo{author}{E.~Mitchell}, \bibinfo{author}{Y.~Lee},
  \bibinfo{author}{A.~Khazatsky}, \bibinfo{author}{C.~D. Manning},
  \bibinfo{author}{C.~Finn},
\newblock \bibinfo{title}{Detectgpt: Zero-shot machine-generated text detection
  using probability curvature},
\newblock \bibinfo{journal}{arXiv preprint arXiv:2301.11305}
  (\bibinfo{year}{2023}).
\bibitem[{Li et~al.(2022)Li, Weng, Song, and Deng}]{li2022artificial}
\bibinfo{author}{B.~Li}, \bibinfo{author}{Y.~Weng}, \bibinfo{author}{Q.~Song},
  \bibinfo{author}{H.~Deng},
\newblock \bibinfo{title}{Artificial text detection with multiple training
  strategies},
\newblock \bibinfo{journal}{arXiv preprint arXiv:2212.05194}
  (\bibinfo{year}{2022}).
\bibitem[{Maloyan et~al.(2022)Maloyan, Nutfullin, and
  Ilyushin}]{maloyan2022dialog}
\bibinfo{author}{N.~Maloyan}, \bibinfo{author}{B.~Nutfullin},
  \bibinfo{author}{E.~Ilyushin},
\newblock \bibinfo{title}{Dialog-22 ruatd generated text detection},
\newblock \bibinfo{journal}{arXiv preprint arXiv:2206.08029}
  (\bibinfo{year}{2022}).
\bibitem[{Mitrovi{\'c} et~al.(2023)Mitrovi{\'c}, Andreoletti, and
  Ayoub}]{mitrovic2023chatgpt}
\bibinfo{author}{S.~Mitrovi{\'c}}, \bibinfo{author}{D.~Andreoletti},
  \bibinfo{author}{O.~Ayoub},
\newblock \bibinfo{title}{Chatgpt or human? detect and explain. explaining
  decisions of machine learning model for detecting short chatgpt-generated
  text},
\newblock \bibinfo{journal}{arXiv preprint arXiv:2301.13852}
  (\bibinfo{year}{2023}).
\bibitem[{Gehrmann et~al.(2019)Gehrmann, Strobelt, and Rush}]{gehrmann2019gltr}
\bibinfo{author}{S.~Gehrmann}, \bibinfo{author}{H.~Strobelt},
  \bibinfo{author}{A.~M. Rush},
\newblock \bibinfo{title}{Gltr: Statistical detection and visualization of
  generated text},
\newblock \bibinfo{journal}{arXiv preprint arXiv:1906.04043}
  (\bibinfo{year}{2019}).
\bibitem[{Gall{\'e} et~al.(2021)Gall{\'e}, Rozen, Kruszewski, and
  Elsahar}]{galle2021unsupervised}
\bibinfo{author}{M.~Gall{\'e}}, \bibinfo{author}{J.~Rozen},
  \bibinfo{author}{G.~Kruszewski}, \bibinfo{author}{H.~Elsahar},
\newblock \bibinfo{title}{Unsupervised and distributional detection of
  machine-generated text},
\newblock \bibinfo{journal}{arXiv preprint arXiv:2111.02878}
  (\bibinfo{year}{2021}).
\bibitem[{Sarvazyan et~al.(2023)Sarvazyan, Gonz{\'a}lez, Franco~Salvador,
  Rangel, Chulvi, and Rosso}]{autextification}
\bibinfo{author}{A.~M. Sarvazyan}, \bibinfo{author}{J.~{\'A}. Gonz{\'a}lez},
  \bibinfo{author}{M.~Franco~Salvador}, \bibinfo{author}{F.~Rangel},
  \bibinfo{author}{B.~Chulvi}, \bibinfo{author}{P.~Rosso},
\newblock \bibinfo{title}{Overview of autextification at iberlef 2023:
  Detection and attribution of machine-generated text in multiple domains},
\newblock in: \bibinfo{booktitle}{Procesamiento del Lenguaje Natural},
  \bibinfo{address}{Jaén, Spain}, \bibinfo{year}{2023}.
\bibitem[{Jim{\'e}nez-Zafra et~al.(2023)Jim{\'e}nez-Zafra, Rangel, and Montes-y
  G{\'o}mez}]{overviewIberLEF2023}
\bibinfo{author}{S.~M. Jim{\'e}nez-Zafra}, \bibinfo{author}{F.~Rangel},
  \bibinfo{author}{M.~Montes-y G{\'o}mez},
\newblock \bibinfo{title}{{Overview of IberLEF 2023: Natural Language
  Processing Challenges for Spanish and other Iberian Languages}},
\newblock \bibinfo{journal}{Procesamiento del Lenguaje Natural}
  \bibinfo{volume}{71} (\bibinfo{year}{2023}).
\bibitem[{Wolf et~al.(2020)Wolf, Debut, Sanh, Chaumond, Delangue, Moi, Cistac,
  Rault, Louf, Funtowicz et~al.}]{wolf2020transformers}
\bibinfo{author}{T.~Wolf}, \bibinfo{author}{L.~Debut},
  \bibinfo{author}{V.~Sanh}, \bibinfo{author}{J.~Chaumond},
  \bibinfo{author}{C.~Delangue}, \bibinfo{author}{A.~Moi},
  \bibinfo{author}{P.~Cistac}, \bibinfo{author}{T.~Rault},
  \bibinfo{author}{R.~Louf}, \bibinfo{author}{M.~Funtowicz}, et~al.,
\newblock \bibinfo{title}{Transformers: State-of-the-art natural language
  processing},
\newblock in: \bibinfo{booktitle}{Proceedings of the 2020 conference on
  empirical methods in natural language processing: system demonstrations},
  \bibinfo{year}{2020}, pp. \bibinfo{pages}{38--45}.
\bibitem[{Sulaiman et~al.(2020)Sulaiman, Wahid, Ariffin, and
  Zulkifli}]{sulaiman2020question}
\bibinfo{author}{S.~Sulaiman}, \bibinfo{author}{R.~A. Wahid},
  \bibinfo{author}{A.~H. Ariffin}, \bibinfo{author}{C.~Z. Zulkifli},
\newblock \bibinfo{title}{Question classification based on cognitive levels
  using linear svc},
\newblock \bibinfo{journal}{Test Eng Manag} \bibinfo{volume}{83}
  (\bibinfo{year}{2020}) \bibinfo{pages}{6463--6470}.
\bibitem[{Liu et~al.(2023)Liu, Gao, Xu, Feng, Ye, Liong, and
  Chen}]{liu2023novel}
\bibinfo{author}{K.-H. Liu}, \bibinfo{author}{J.~Gao}, \bibinfo{author}{Y.~Xu},
  \bibinfo{author}{K.-J. Feng}, \bibinfo{author}{X.-N. Ye},
  \bibinfo{author}{S.-T. Liong}, \bibinfo{author}{L.-Y. Chen},
\newblock \bibinfo{title}{A novel soft-coded error-correcting output codes
  algorithm},
\newblock \bibinfo{journal}{Pattern Recognition} \bibinfo{volume}{134}
  (\bibinfo{year}{2023}) \bibinfo{pages}{109122}.
\bibitem[{Hong and Cho(2008)}]{hong2008probabilistic}
\bibinfo{author}{J.-H. Hong}, \bibinfo{author}{S.-B. Cho},
\newblock \bibinfo{title}{A probabilistic multi-class strategy of one-vs.-rest
  support vector machines for cancer classification},
\newblock \bibinfo{journal}{Neurocomputing} \bibinfo{volume}{71}
  (\bibinfo{year}{2008}) \bibinfo{pages}{3275--3281}.
\bibitem[{Mahabub(2020)}]{mahabub2020robust}
\bibinfo{author}{A.~Mahabub},
\newblock \bibinfo{title}{A robust technique of fake news detection using
  ensemble voting classifier and comparison with other classifiers},
\newblock \bibinfo{journal}{SN Applied Sciences} \bibinfo{volume}{2}
  (\bibinfo{year}{2020}) \bibinfo{pages}{525}.
\bibitem[{He et~al.(2021)He, Liu, Gao, and Chen}]{he2021deberta}
\bibinfo{author}{P.~He}, \bibinfo{author}{X.~Liu}, \bibinfo{author}{J.~Gao},
  \bibinfo{author}{W.~Chen},
\newblock \bibinfo{title}{Deberta: Decoding-enhanced bert with disentangled
  attention},
\newblock in: \bibinfo{booktitle}{International Conference on Learning
  Representations}, \bibinfo{year}{2021}. \URLprefix
  \url{https://openreview.net/forum?id=XPZIaotutsD}.
\bibitem[{Reimers and Gurevych(2019)}]{reimers-2019-sentence-bert}
\bibinfo{author}{N.~Reimers}, \bibinfo{author}{I.~Gurevych},
\newblock \bibinfo{title}{Sentence-bert: Sentence embeddings using siamese
  bert-networks},
\newblock in: \bibinfo{booktitle}{Proceedings of the 2019 Conference on
  Empirical Methods in Natural Language Processing},
  \bibinfo{publisher}{Association for Computational Linguistics},
  \bibinfo{year}{2019}. \URLprefix \url{http://arxiv.org/abs/1908.10084}.
\bibitem[{Solaiman et~al.(2019)Solaiman, Brundage, Clark, Askell, Herbert-Voss,
  Wu, Radford, Krueger, Kim, Kreps et~al.}]{solaiman2019release}
\bibinfo{author}{I.~Solaiman}, \bibinfo{author}{M.~Brundage},
  \bibinfo{author}{J.~Clark}, \bibinfo{author}{A.~Askell},
  \bibinfo{author}{A.~Herbert-Voss}, \bibinfo{author}{J.~Wu},
  \bibinfo{author}{A.~Radford}, \bibinfo{author}{G.~Krueger},
  \bibinfo{author}{J.~W. Kim}, \bibinfo{author}{S.~Kreps}, et~al.,
\newblock \bibinfo{title}{Release strategies and the social impacts of language
  models},
\newblock \bibinfo{journal}{arXiv preprint arXiv:1908.09203}
  (\bibinfo{year}{2019}).
\bibitem[{la~Rosa y Eduardo G. Ponferrada y Manu Romero y Paulo Villegas y
  Pablo González de Prado Salas~y María~Grandury(2022)}]{BERTIN}
\bibinfo{author}{J.~D. la~Rosa y Eduardo G. Ponferrada y Manu Romero y Paulo
  Villegas y Pablo González de Prado Salas~y María~Grandury},
\newblock \bibinfo{title}{Bertin: Efficient pre-training of a spanish language
  model using perplexity sampling},
\newblock \bibinfo{journal}{Procesamiento del Lenguaje Natural}
  \bibinfo{volume}{68} (\bibinfo{year}{2022}) \bibinfo{pages}{13--23}.
  \URLprefix
  \url{http://journal.sepln.org/sepln/ojs/ojs/index.php/pln/article/view/6403}.
\bibitem[{Fandiño et~al.(2022)Fandiño, Estapé, Pàmies, Palao, Ocampo,
  Carrino, Oller, Penagos, Agirre, and Villegas}]{MARIA}
\bibinfo{author}{A.~G. Fandiño}, \bibinfo{author}{J.~A. Estapé},
  \bibinfo{author}{M.~Pàmies}, \bibinfo{author}{J.~L. Palao},
  \bibinfo{author}{J.~S. Ocampo}, \bibinfo{author}{C.~P. Carrino},
  \bibinfo{author}{C.~A. Oller}, \bibinfo{author}{C.~R. Penagos},
  \bibinfo{author}{A.~G. Agirre}, \bibinfo{author}{M.~Villegas},
\newblock \bibinfo{title}{Maria: Spanish language models},
\newblock \bibinfo{journal}{Procesamiento del Lenguaje Natural}
  \bibinfo{volume}{68} (\bibinfo{year}{2022}). \URLprefix
  \url{https://upcommons.upc.edu/handle/2117/367156#.YyMTB4X9A-0.mendeley}.
  \DOIprefix\doi{10.26342/2022-68-3}.
\bibitem[{Conneau et~al.(2019)Conneau, Khandelwal, Goyal, Chaudhary, Wenzek,
  Guzm{\'a}n, Grave, Ott, Zettlemoyer, and Stoyanov}]{conneau2019unsupervised}
\bibinfo{author}{A.~Conneau}, \bibinfo{author}{K.~Khandelwal},
  \bibinfo{author}{N.~Goyal}, \bibinfo{author}{V.~Chaudhary},
  \bibinfo{author}{G.~Wenzek}, \bibinfo{author}{F.~Guzm{\'a}n},
  \bibinfo{author}{E.~Grave}, \bibinfo{author}{M.~Ott},
  \bibinfo{author}{L.~Zettlemoyer}, \bibinfo{author}{V.~Stoyanov},
\newblock \bibinfo{title}{Unsupervised cross-lingual representation learning at
  scale},
\newblock \bibinfo{journal}{arXiv preprint arXiv:1911.02116}
  (\bibinfo{year}{2019}).
\bibitem[{Beltagy et~al.(2019)Beltagy, Lo, and
  Cohan}]{beltagy-etal-2019-scibert}
\bibinfo{author}{I.~Beltagy}, \bibinfo{author}{K.~Lo},
  \bibinfo{author}{A.~Cohan},
\newblock \bibinfo{title}{Scibert: A pretrained language model for scientific
  text},
\newblock in: \bibinfo{booktitle}{EMNLP}, \bibinfo{publisher}{Association for
  Computational Linguistics}, \bibinfo{year}{2019}. \URLprefix
  \url{https://www.aclweb.org/anthology/D19-1371}.
\bibitem[{Liu et~al.(2019)Liu, Ott, Goyal, Du, Joshi, Chen, Levy, Lewis,
  Zettlemoyer, and Stoyanov}]{DBLP:journals/corr/abs-1907-11692}
\bibinfo{author}{Y.~Liu}, \bibinfo{author}{M.~Ott}, \bibinfo{author}{N.~Goyal},
  \bibinfo{author}{J.~Du}, \bibinfo{author}{M.~Joshi},
  \bibinfo{author}{D.~Chen}, \bibinfo{author}{O.~Levy},
  \bibinfo{author}{M.~Lewis}, \bibinfo{author}{L.~Zettlemoyer},
  \bibinfo{author}{V.~Stoyanov},
\newblock \bibinfo{title}{Roberta: {A} robustly optimized {BERT} pretraining
  approach},
\newblock \bibinfo{journal}{CoRR} \bibinfo{volume}{abs/1907.11692}
  (\bibinfo{year}{2019}). \URLprefix \url{http://arxiv.org/abs/1907.11692}.
  \href{http://arxiv.org/abs/1907.11692}{{\tt arXiv:1907.11692}}.
\bibitem[{Beltagy et~al.(2020)Beltagy, Peters, and
  Cohan}]{Beltagy2020Longformer}
\bibinfo{author}{I.~Beltagy}, \bibinfo{author}{M.~E. Peters},
  \bibinfo{author}{A.~Cohan},
\newblock \bibinfo{title}{Longformer: The long-document transformer},
\newblock \bibinfo{journal}{arXiv:2004.05150}  (\bibinfo{year}{2020}).
\bibitem[{Devlin et~al.(2018)Devlin, Chang, Lee, and
  Toutanova}]{DBLP:journals/corr/abs-1810-04805}
\bibinfo{author}{J.~Devlin}, \bibinfo{author}{M.~Chang},
  \bibinfo{author}{K.~Lee}, \bibinfo{author}{K.~Toutanova},
\newblock \bibinfo{title}{{BERT:} pre-training of deep bidirectional
  transformers for language understanding},
\newblock \bibinfo{journal}{CoRR} \bibinfo{volume}{abs/1810.04805}
  (\bibinfo{year}{2018}). \URLprefix \url{http://arxiv.org/abs/1810.04805}.
  \href{http://arxiv.org/abs/1810.04805}{{\tt arXiv:1810.04805}}.

\end{thebibliography}

\end{document}